\documentclass[conference]{IEEEtran}
\PassOptionsToPackage{numbers,compress}{natbib}
\usepackage[numbers]{natbib}
\usepackage[title]{appendix}
\usepackage{multicol}
\usepackage{bm}
\usepackage{times}

\usepackage{graphicx}
\usepackage{amsmath}
\usepackage{amssymb}
\usepackage{booktabs}
\usepackage{cuted}
\usepackage{duckuments}
\usepackage{subcaption}

\usepackage{appendix}
\usepackage{titletoc}

\usepackage[pagebackref,breaklinks,colorlinks]{hyperref}

\usepackage[capitalize]{cleveref}
\Crefname{section}{Sec.}{Secs.}
\crefname{section}{Sec.}{Secs.}
\Crefname{table}{Tab.}{Tabs.}
\crefname{table}{Tab.}{Tabs.}
\Crefname{figure}{Fig.}{Figs.}
\crefname{figure}{Fig.}{Figs.}

\usepackage[utf8]{inputenc} 
\usepackage[T1]{fontenc}    
\usepackage{url}            
\usepackage{amsfonts}       
\usepackage{nicefrac}       
\usepackage[nopatch=eqnum]{microtype}      
\usepackage{xcolor}         

\usepackage{epsfig}
\usepackage{tabularx}
\usepackage{bbm}
\usepackage{color}
\usepackage{wrapfig}
\usepackage{subcaption}
\usepackage{float}
\usepackage{makecell}
\usepackage[percent]{overpic}
\usepackage{adjustbox}

\usepackage{multirow}
\usepackage{inconsolata}
\usepackage{comment}
\usepackage{xspace}
\usepackage{utfsym}

\usepackage[font=small]{caption}

\newcommand\mypar[1]
{\par\vspace{1.0mm}\noindent\textbf{#1}\;\;}

\newcommand{\GG}{\mathcal{G}}
\newcommand{\VV}{\mathcal{V}}
\newcommand{\EE}{\mathcal{E}}
\newcommand{\LL}{\mathcal{L}}

\newcolumntype{Y}{>{\centering\arraybackslash}X}

\makeatletter
\def\blfootnote{\gdef\@thefnmark{}\@footnotetext}
\makeatother

\usepackage{hyperref}
\newcounter{mysection}

\begin{document}

\title{AdaptiGraph: Material-Adaptive Graph-Based \\ Neural Dynamics for Robotic Manipulation}

\author{Kaifeng Zhang$^1$\textsuperscript{*}, 
        Baoyu Li$^1$\textsuperscript{*},  
        Kris Hauser$^1$,
        Yunzhu Li$^{1,2}$ \\
        $^1$University of Illinois Urbana-Champaign \quad $^2$Columbia University
        }

\maketitle
\IEEEpeerreviewmaketitle
\blfootnote{* Denotes equal contribution.}

\setcounter{figure}{0}
\setcounter{table}{0}
\setcounter{section}{0}
\renewcommand{\thesection}{\Roman{section}}
\renewcommand*{\theHsection}{\thesection} 

\begin{strip}
    \centering
    \vspace{-40pt}
    \includegraphics[width=\linewidth]{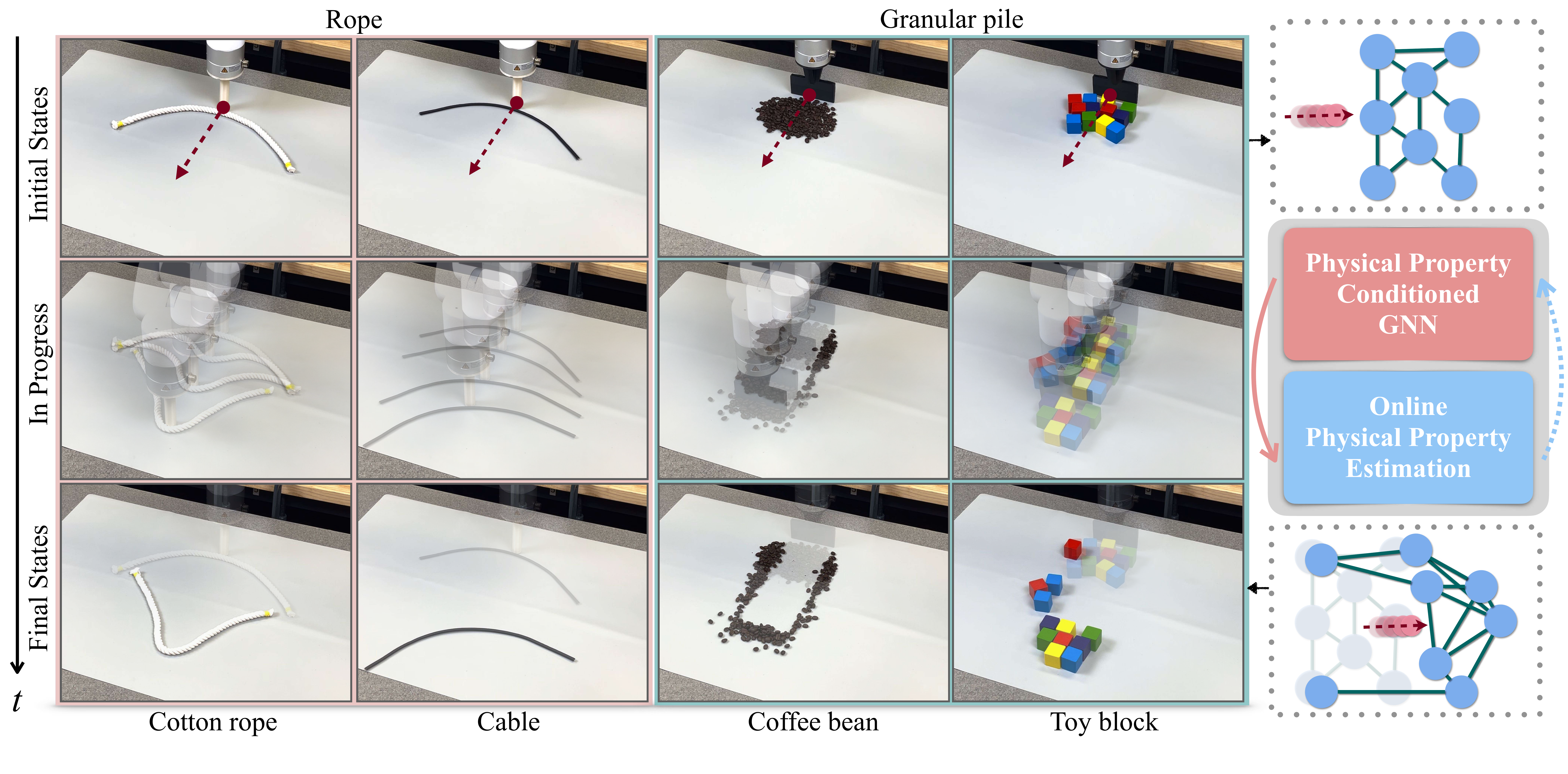}
    \vspace{-28pt}
    \captionof{figure}{\small
    \textbf{Motivation.} Objects made from different materials can exhibit distinct behaviors under interaction. Even within the same object category, varying physical parameters like stiffness can lead to different behaviors. Examples shown here include handling cotton rope and cable, as well as arranging granular piles such as coffee beans and toy blocks. Although the initial configuration and action are the same, different physical parameters result in distinct final states, necessitating the need for online adaptation for effective manipulation.
    To this end, we introduce \textbf{AdaptiGraph}, a unified graph-based neural dynamics framework for real-time modeling and control of various materials with unknown physical properties. \textbf{AdaptiGraph} integrates a \textbf{physical property-conditioned dynamics model} with \textbf{online physical property estimation}. Our framework enables robots to adaptively manipulate diverse objects with varying physical properties and dynamics.
    }
    \label{fig:teaser}
    \vspace{-10pt}
\end{strip}

\begin{abstract}

Predictive models are a crucial component of many robotic systems. Yet, constructing accurate predictive models for a variety of deformable objects, especially those with unknown physical properties, remains a significant challenge. This paper introduces AdaptiGraph, a learning-based dynamics modeling approach that enables robots to predict, adapt to, and control a wide array of challenging deformable materials with unknown physical properties. AdaptiGraph leverages the highly flexible graph-based neural dynamics (GBND) framework, which represents material bits as particles and employs a graph neural network (GNN) to predict particle motion. Its key innovation is a unified physical property-conditioned GBND model capable of predicting the motions of diverse materials with varying physical properties without retraining. Upon encountering new materials during online deployment, AdaptiGraph utilizes a physical property optimization process for a few-shot adaptation of the model, enhancing its fit to the observed interaction data. The adapted models can precisely simulate the dynamics and predict the motion of various deformable materials, such as ropes, granular media, rigid boxes, and cloth, while adapting to different physical properties, including stiffness, granular size, and center of pressure. On prediction and manipulation tasks involving a diverse set of real-world deformable objects, our method exhibits superior prediction accuracy and task proficiency over non-material-conditioned and non-adaptive models. The project page is available at \texttt{\url{https://robopil.github.io/adaptigraph/}}.

\end{abstract}

\section{Introduction}
\label{sec:intro}

Learning predictive models, also known as system identification, is a crucial component of many robotic tasks. Whereas classical methods rely on the explicit parameterization of the system state and struggle with systems that have high degrees of freedom, a significant body of work over the last decade has attempted to learn models directly from visual observations. Prior approaches have learned predictive models based on pixels~\cite{finn2017deep,hoque2020visuospatial} or latent representations of images~\cite{hafner2019dream, hafner2020mastering}. However, such representations often overlook the structure of the environment and do not generalize well across different camera poses, object poses, robots, object sizes, and object shapes. Recently, a series of studies have employed Graph Neural Networks (GNN) to model environments as 3D particles and their pairwise interactions~\cite{li2019propagation, shi2022robocraft, shi2023robocook, Wang-RSS-23}. A graph representation has proven effective in capturing relational bias and predicting complex motions of deformable objects, but prior works typically only focus on a single material and would require extensive training to model an object of new material or with unknown physical properties. Hence, it is an important challenge to provide such graph-based models to adapt to objects and tasks involving diverse materials and varying physical properties, such as manipulating ropes with different stiffness and granular media with different granularity~(Fig.~\ref{fig:teaser}).

In this work, we present a unified framework for modeling the dynamics of objects with different materials and physical properties. In addition to classifying objects into discrete material types such as rigid objects, ropes, etc., we further consider a range of intra-class physical property variations in each material type. We propose to encode this variation using a continuous variable which we call the physical property variable, and integrate the variable into a Graph-Based Neural Dynamics (GBND) framework (Fig.~\ref{fig:teaser}). The physical property variable indicates the important intrinsic properties of each material category, including stiffness for deformable objects and the center of pressure position for rigid objects. By encoding the material type and physical property variables into particles in the graph, the model learns material-specific dynamic functions that predict different physical behaviors for objects with different physical properties. We then employ a test-time adaptation method to reason about the physical properties of novel objects. Specifically, the robot actively interacts with the novel object, observes its response, and estimates its physical properties to optimize the model's fit to the observed reactions. The estimation is performed in a few-shot manner and can be directly applied to planning and trajectory optimization for downstream manipulation tasks.

In our experiments, we verify this framework on four types of objects: rigid objects, granular objects, rope-like objects, and cloth-like objects. Experiments show that our framework can distinguish and model the dynamics of objects across a broad range of physical properties, for instance, from very soft ropes like yarn and shoelaces to very stiff ropes like cables, and from very fine-grained granular matter like coffee beans to very coarse-grained ones like toy blocks (Fig.~\ref{fig:teaser}). The model is trained on diverse data collected with a simulator and tested with online adaptation on real objects. The results demonstrate that (1) our adaptation module provides consistent and interpretable estimates of the objects' physical property variables, and (2) by conditioning on the estimated physical property variable, the model can carry out more accurate dynamics estimation and more efficient manipulation, especially for objects with extreme or out-of-distribution physical properties.

\section{Related Work}

\subsection{Model Learning for Robotic Manipulation}
\label{sec:model}
Analytical physics-based models facilitate a wide span of robotic manipulation tasks~\cite{hogan2016feedback, pang2023global,yu2016more}. However, building accurate physics models is often infeasible in the real world due to unobservable physics properties such as mass, friction, and stiffness, occluded surfaces of geometry, sensitivity to parameter estimates, and the high computational expense of simulating deformable objects. To mitigate these issues, recent approaches apply learning-based techniques to obtain dynamics models directly from sensory inputs~\cite{chua2018deep, PDDM, yang2023tacgnnlearning, hoque2020visuospatial, finn2017deep, babaeizadeh2017stochastic}. Graph-based representations and GNNs have been proven effective in modeling the complex behaviors of non-rigid objects due to their ability to capture spatial relational bias~\cite{battaglia2018relational, pfaff2020learning, li2019propagation, lin2022learning, sanchez2020learning, Wang-RSS-23}. 
Prior work has explored the application of graph-based dynamics models on a variety of material types, including rigid bodies~\cite{li2019propagation, huang2023defgraspnets, liu2023modelbased}, plasticine~\cite{shi2022robocraft, shi2023robocook}, clothes~\cite{puthuveetil2023robust, lin2022learning, pfaff2020learning, longhini2023edo}, fluids~\cite{li2018learning, sanchez2020learning}, and granular matter~\cite{Wang-RSS-23}. However, nearly all of these approaches focus on a single type of material and fail to consider variation in physical properties, thus limiting their generalization and adaptation capabilities. In contrast, our method considers a wider range of materials and variations in physical properties in a single property-conditioned graph-based neural dynamics models, and this enables our approach to adaptively estimate the unknown physical properties of unseen objects through interaction. 

\subsection{Physical Property Estimation and Few-Shot Adaptation}
\label{sec:related_physics}

Estimating and adapting to the physical properties of unseen objects is an inherent challenge in various robotic applications. Previous works have attempted to infer physical properties by tweaking parameters in physics-based simulations~\cite{ding2021dynamic, tung2023physion++, NEURIPS2019_28f0b864, frank2010learning, sundaresan2022diffcloud, galileo} or utilizing extra modalities, e.g., tactile signal~\cite{she2020cable, wang2020swingbot, yao2023icra}, but these approaches have a high demand for the object's full state information, or require extra sensors. In comparison, adaptively learning explicit physical property variables~\cite{agrawal2016learning, li2018learning, DensePhysNet, chen2022comphy, li2020visual, longhini2023edo} or low-dimensional latent representations that implicitly encode physical properties~\cite{kumar2021rma, evans2022context} in a neural network only requires partial observations and few-shot exploratory interactions as input. A line of work goes further by using the large vision-language models to infer physical properties solely from static observation~\cite{gao2023physically, wang2023newton}, but these estimations are rough and do not involve actual interactions. For estimation/adaptation from interactions, previous efforts were still limited to simulations or focused only on single types of materials, e.g., rigid objects. There is also a dilemma in choosing the representation form: explicit variables suffer from domain gaps such as the sim-to-real gap, yet latent representation has relatively lower interpretability. In contrast, our approach incorporates a graph-structured model within an inverse optimization framework, offering interpretability, generalizability to objects beyond the training distribution, and applicability to a broader array of material types, including rigid boxes, ropes, cloths, and granular substances, in the real-world scenario.

\begin{figure*}[t]
    \centering
    \includegraphics[width=\linewidth]{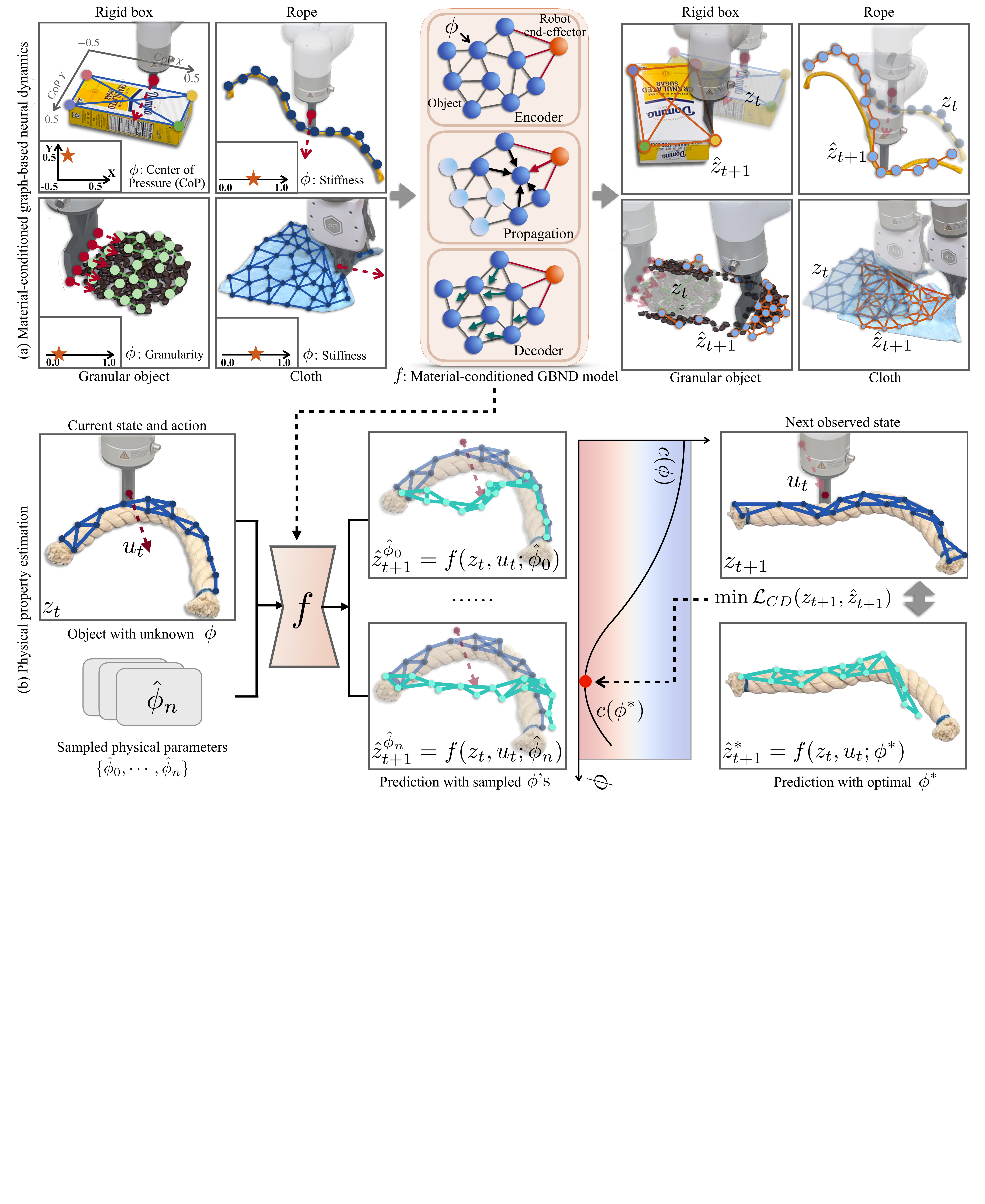}
    \vspace{-220pt}
    \caption{\small
    \textbf{Overview of proposed framework: AdaptiGraph.} \textbf{(a)} Our graph-based dynamics model $f$ is conditioned on the discrete material type and continuous physical parameters $\phi$. $\phi$ is encoded as the node features, which will be propagated and updated in the model training process. Our model can accurately predict the future state $\hat{z}_{t+1}$ for a variety of objects with different physical properties. \textbf{(b)} Our framework performs physical property estimation for few-shot adaptation. This is achieved through an inverse optimization process to estimate the optimal physical parameters as predicted by the learned dynamics model $f$. The optimal physical parameter $\phi^*$ is identified by minimizing the cost function, which is defined as the Chamfer Distance between the predicted graph state and the actual future graph state.
    }
    \label{fig:overview}
    \vspace{-15pt}
\end{figure*}

\section{Methods}

We first introduce the problem formulation in Sec.~\ref{sec:formulation}. Then, we introduce the perception module and the structure of our physical property-conditioned graph-based dynamics model in Sec.~\ref{sec:dynamics}. We discuss the test-time adaptation algorithm for physics property estimation in Sec.~\ref{sec:adaptation}. Finally, in Sec.~\ref{sec:control}, we introduce how we perform closed-loop control for downstream manipulation tasks. 

\subsection{Problem Formulation}
\label{sec:formulation}
Our aim is to learn a dynamics model, $f$, that is conditioned on the material type $M$ and continuous physical property variable $\phi$, and develop a test-time few-shot adaptation scheme to infer the physical property variable for unseen objects. 
Specifically, the dynamics model predicts how the environment will change if the robot applies a given action:
\begin{equation}
\hat{z}_{t+1} = f(z_t, u_t; \phi, M),
\end{equation}
where $M$ indicates the material type (e.g., rigid, granular, rope, cloth), $\phi$ indicates material-specific physical property variables, and $u_t$, $z_t$, $z_{t+1}$ are the robot action, current environment state at time $t$, and the next state at time $t + 1$, respectively. In our approach, we train the dynamics model to minimize the accumulated future prediction loss.

By conditioning on $M$ and $\phi$, the model learns to predict material-dependent physical behaviors, based on which we can perform physical property estimation through the following optimization problem: 
\begin{equation}
\label{eq:1}
\phi^* = \arg \min_{\phi} \sum_{t=1}^T \text{cost}(\hat{z}_{t+1}, z_{t+1}),
\end{equation}
where $T$ is the iteration number indicating the number of interactions with the unseen object, and $\text{cost}(\cdot, \cdot)$ is the cost function measuring the discrepancy between the predicted future state $\hat{z}_{t+1}$ and the observed state $z_{t+1}$.

\subsection{Material-Conditioned Graph-based Neural Dynamics Model}
\label{sec:dynamics}

We propose to instantiate the dynamics model with a graph neural network.  Following prior work on graph-based neural dynamics (GBND)~\cite{li2018learning, Wang-RSS-23, shi2022robocraft, shi2023robocook}, we define the environment state as a graph: $z_t \overset{\underset{\mathrm{\Delta }}{}}{=} \GG_t =  ( \VV_t, \EE_t )$, where $\VV_t$ is the vertex set representing object particles, and $\EE_t$ denotes the edge set representing interacting particle pairs at time step $t$. Given the point cloud input, the object particle positions are determined by the farthest point sampling method~\cite{moenning2003fast}, which ensures sufficient coverage of the object's geometry. We construct edges between particles based on a spatial distance threshold $d$. We also sample particles on the robot end-effector and construct relations between robot particles and object particles. 

The main improvement of our model over previous works based on GBND is our material- and physical property-conditioning module (Fig.~\ref{fig:overview}a). Suppose a vertex $v_{i, t} \in \VV_t$ has material $M_i$ and physical property variable $\phi_i$.
We incorporate this material information into the vertex features along with the 3D position information over $h$ history timesteps $x_{i,t-h:t}$ and the vertex attribute $c^v_{i,t}$ which indicates whether the particle belongs to an object or the robot end-effector. Formally, $v_{i,t} = (x_{i,t-h:t}, c^v_{i,t}, \phi_i, M_i) \in \VV_t$. The history positions implicitly encode the velocity information. Empirically, we choose $h=3$.
The relation features between a pair of particles is denoted as $e_{k,t} = (w, u, c^e_{k,t}) \in \EE_t$, where $1 \leq w, u \leq |\VV_t|$ are the receiver particle index and the sender particle index of the $k^{th}$ edge respectively. The edge attribute $c^e_{k,t}$ contains information such as whether the sender and receiver belong to the same or different objects. 

The constructed vertex and edge features are first fed into the vertex encoder $f_\VV^{enc}$ and the edge encoder $f_\EE^{enc}$ respectively to get the latent vertex and edge embeddings $h_{v_i, t}$ and $h_{e_k, t}$:
\begin{align}
    h^0_{v_i, t} = f_{\VV}^{enc}(v_{i, t}), \quad
    h^0_{e_k, t} = f_{\EE}^{enc}(v_{w,t}, v_{u, t}).
\end{align}
Then, an edge propagation network $f_{\EE}^{prop}$ and vertex propagation network $f_{\VV}^{prop}$ performs iterative update of the vertex and edge embeddings to perform multi-step message passing. Specifically, for $l = 0, 1, 2, \cdots, L-1$, a single message passing step is as follows:
\begin{align}
    h^{l+1}_{e_k, t} &= f_{\EE}^{prop}(h^l_{w, t}, h^l_{u, t}), \\
    h^{l+1}_{v_i, t} &= f_{\VV}^{prop}(h^l_{v_i, t}, \sum_{j\in \mathcal{N}(v_{i,t})} h^{l+1}_{e_j, t}),
\end{align}
where $\mathcal{N}(v_{i,t})$ indicates the index set of edges in which vertex $i$ is the receiver at time $t$, and $L$ is the total number of message passing steps. Finally, one vertex decoder $f_{\VV}^{dec}$ predicts the system's state at the next time step: $\hat{v}_{i, t+1} = f_{\VV}^{dec} (h^L_{v_i,t})$.

\textit{Translation equivariance.} Translation equivariance is a desired property for dynamics models. Formally, for any global 3D translation added to the particle locations, the predictions should also be translated identically. We enforce translation equivariance by passing the position difference of receiver and sender particles to the edge encoder $f_\EE^{enc}$, instead of passing absolution particle positions to the vertex encoder $f_\VV^{enc}$.

\textit{Training.} To regulate the cumulative dynamics prediction error, we supervise the model's prediction results on $K$ prediction steps and perform backpropagation through time to optimize model parameters. In practice, we choose $K=3$ for all tasks for balancing efficiency and performance. We use MSE loss on predicted object particle positions as the loss function:
\begin{equation}
    \LL = \sum_t ||z_{t+1} - f (z_{t}, u_t; \phi, M)||_2^2,
\end{equation}

To obtain training data at scale, we generate diverse object trajectories by randomizing robot actions and object configurations using physics-based simulators. Most importantly, we randomize the material configuration for each instance in the dataset. To achieve this, we identify the physics property $\phi$ and randomize the property over a wide range of feasible values.  

\subsection{Few-Shot Physical Property Adaptation}
\label{sec:adaptation}

After learning the material-conditioned GBND model, we deploy the model to objects with unknown physical properties in the real world. Inspired by human's ability to reason about objects' physical properties by interacting with them, we design an inverse optimization pipeline through few-shot curiosity-driven interaction.

Specifically, to estimate the physical property variable, the robot actively interacts with the object. In each iteration, it selects the action that maximizes the predicted displacement of the object. Intuitively, the action that maximizes displacement is likely to reveal more information about the object's physical properties than random actions would. 

After each interaction, the robot updates its estimate of the object by minimizing the dynamics prediction error from previous interactions. As the robot undergoes several interactions, the estimation of physical property tends to stabilize, reaching the final optimized value.

In our experiments, we adopt a fixed number of iterations for adaptation. We measure the displacement of the object by computing the Chamfer Distance (CD) between the current state $z_t$ and the predicted state $\hat{z}_t$:
\begin{equation}
    \LL_{CD}(\hat{z}_t, z_t) = \sum_{x\in {\color{black}\VV_t}} \min_{y\in {\color{black}\hat{\VV}_t}} ||x-y||_2^2 + \sum_{y\in {\color{black}\hat{\VV}_t}} \min_{x \in {\color{black}\VV_t}} ||x-y||_2^2,
\end{equation}
where $\hat{\VV}_t$ and $\hat{\VV}_t$ denote the vertex sets at state $z_t$ and $\hat{z}_t$, respectively. The actions for curiosity-driven interactions are optimized using the Model-Predictive Path Integral (MPPI)~\cite{williams2017model} trajectory optimization algorithm to maximize the above Chamfer Distance.

For inverse optimization at the $t^{th}$ interaction step, we adopt gradient-free optimizers including Bayesian Optimization (BO) for single-dimensional physical property variables and CMA-ES for multi-dimensional variables. We instantiate the optimization problem described in Eq.~\ref{eq:1} by specifying the cost function $\text{cost}(\hat{z}_{i+1}, z_{i+1})$ as the Chamfer Distance between the dynamics prediction and the true outcome after each interaction:

\begin{equation}
    \hat{\phi}_{t} = \arg\min_{\phi}\sum_{i=0}^{t-1}  \LL_{CD} (\hat{z}_{i+1}, z_{i+1}),
\end{equation}
where $\hat{z}_{i+1} = f(z_i, u_i; \phi, M)$.

For some materials whose physical properties span a large range (e.g., stiffness for ropes), the test object can potentially fall outside the training distribution of the model. Our material-conditioned model allows for generalization beyond the training domain by directly setting the domain of $\hat{\phi}$ at the adaptation stage to be an extension of the maximal range of $\phi$ in the training data. Specifically, the minimum value and maximum value for $\hat{\phi}$ is $\phi_{min} - 0.2 (\phi_{max} - \phi_{min})$ and $\phi_{max} + 0.2 (\phi_{max} - \phi_{min})$, where $\phi_{max}$ and $\phi_{min}$ are the maximum and minimum value of $\phi$ in the training dataset.

\subsection{Closed-Loop Model-Based Planning}
\label{sec:control}
Using the estimated physics parameter $\hat{\phi}$, the learned model $f$ adapts to new objects, yielding lower dynamics prediction errors on the few-shot, curiosity-driven online interaction dataset. Thus, we can also use the adapted model to perform closed-loop planning for material-specific manipulation tasks within a Model Predictive Control (MPC) framework~\cite{camacho2013model}. The improved dynamics prediction accuracy after few-shot adaptation will help the robot manipulate the object more efficiently and effectively towards goal configurations.

Concretely, the model-based control pipeline is defined as follows: given the state space $\mathbb{Z}$ and the action space $\mathbb{U}$, the cost function is a mapping from $\mathbb{Z} \times \mathbb{U}$ to $\mathbb{R}$. For each starting state $z_0 \in \mathbb{Z}$, we iteratively sample actions $\{u_i\}_{i=1}^T$ in the action space, apply the learned dynamics model to predict the outcome, and apply the MPPI trajectory optimization algorithm for the action sequence $\{u_i\}$ that minimizes the cost function. 
In our experiments, the cost function includes a task-related term that measures the distance from the current state to the desired target, along with other penalty terms for infeasible actions and collision avoidance. Please refer to Sec.~\ref{sec:supp_mppi} of the supplementary material for details.

\section{Experiments}

\begin{figure}[t]
    \centering
    \includegraphics[width=\linewidth]{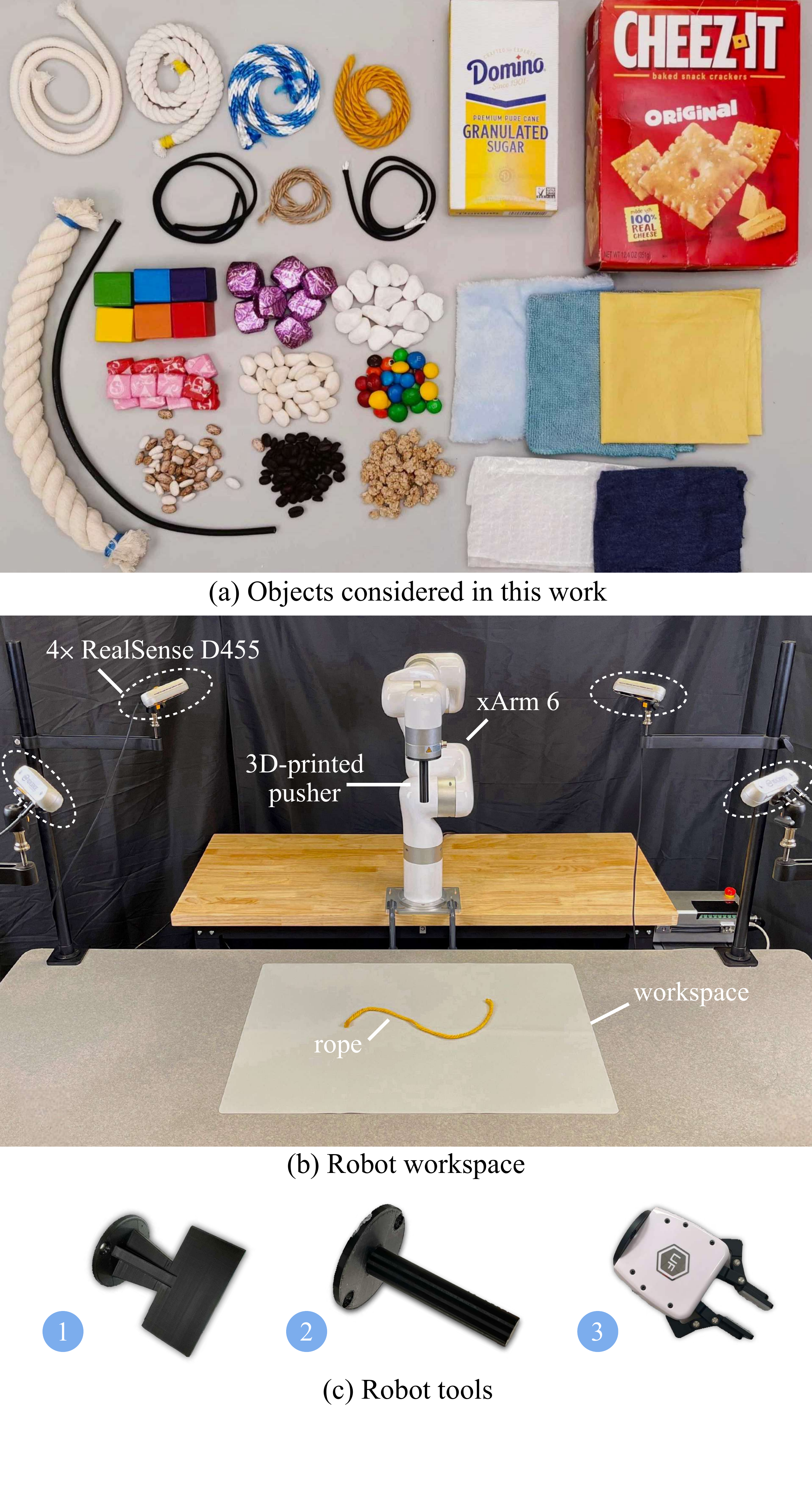}
    \vspace{-35pt}
    \caption{\small
    \textbf{Real-world setup.} \textbf{(a)} Our study involves 22 objects categorized into four types of materials, each with distinct physical characteristics: (i) 9 varieties of ropes, such as cotton ropes and cables, (ii) 9 granular materials, including items like toy blocks and coffee beans, (iii) 5 pieces of cloth made from different fabrics like cotton and synthetic fibers, (iv) 2 boxes of varying shapes, whose centers of pressure we alter by placing weights inside them. \textbf{(b)} The dashed white circles show four calibrated RGB-D cameras mounted at four corners of the table. The robot is outfitted with specialized end effectors to interact with the objects in its operational area. \textbf{(c)} We employ three different tools for specific tasks: (1) a flat pusher for granular piles gathering, (2) a cylindrical pusher for pushing rigid boxes and straightening ropes, (3) an xArm gripper for cloth relocating. 
    }
    \label{fig:setup}
    \vspace{-20pt}
\end{figure}

\begin{figure*}[t]
    \centering
    \vspace{-20pt}
    \includegraphics[width=\linewidth]{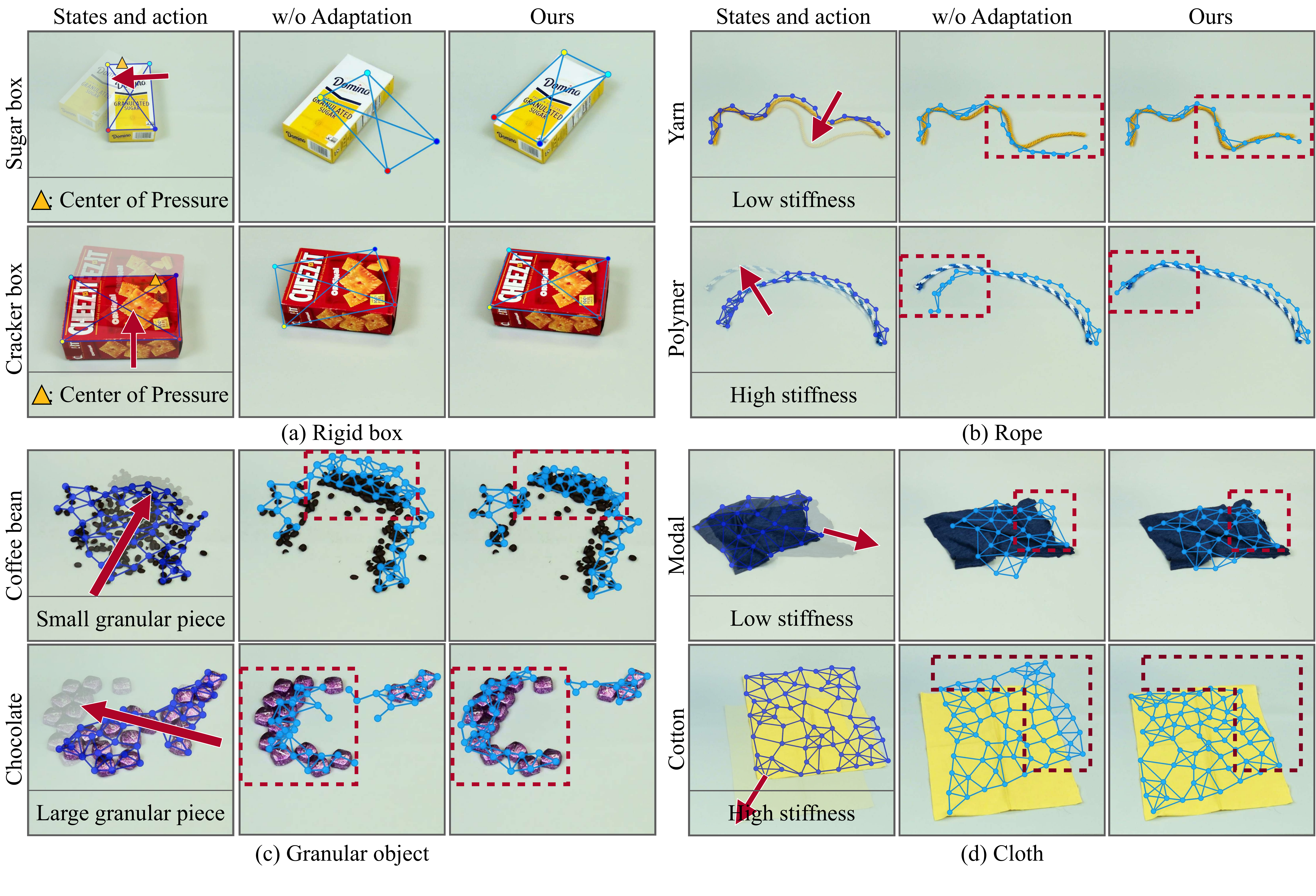}
    \vspace{-27pt}
    \caption{\small
    \textbf{Qualitative results on dynamics prediction.} We conduct qualitative comparisons to assess the performance of our method against the baseline of a GNN without adaptation, focusing on the one-step prediction of dynamics across eight objects within four distinct material categories exhibiting varying extreme physical properties. The results, delineated by red dashed boxes, demonstrate that our approach surpasses the baseline in accurately capturing the variations in dynamics that arise due to differences in the objects' physical properties.
    }
    \label{fig:qual_dynamics}
    \vspace{-18pt}
\end{figure*}

\begin{figure}
    \centering
    \includegraphics[width=\linewidth]{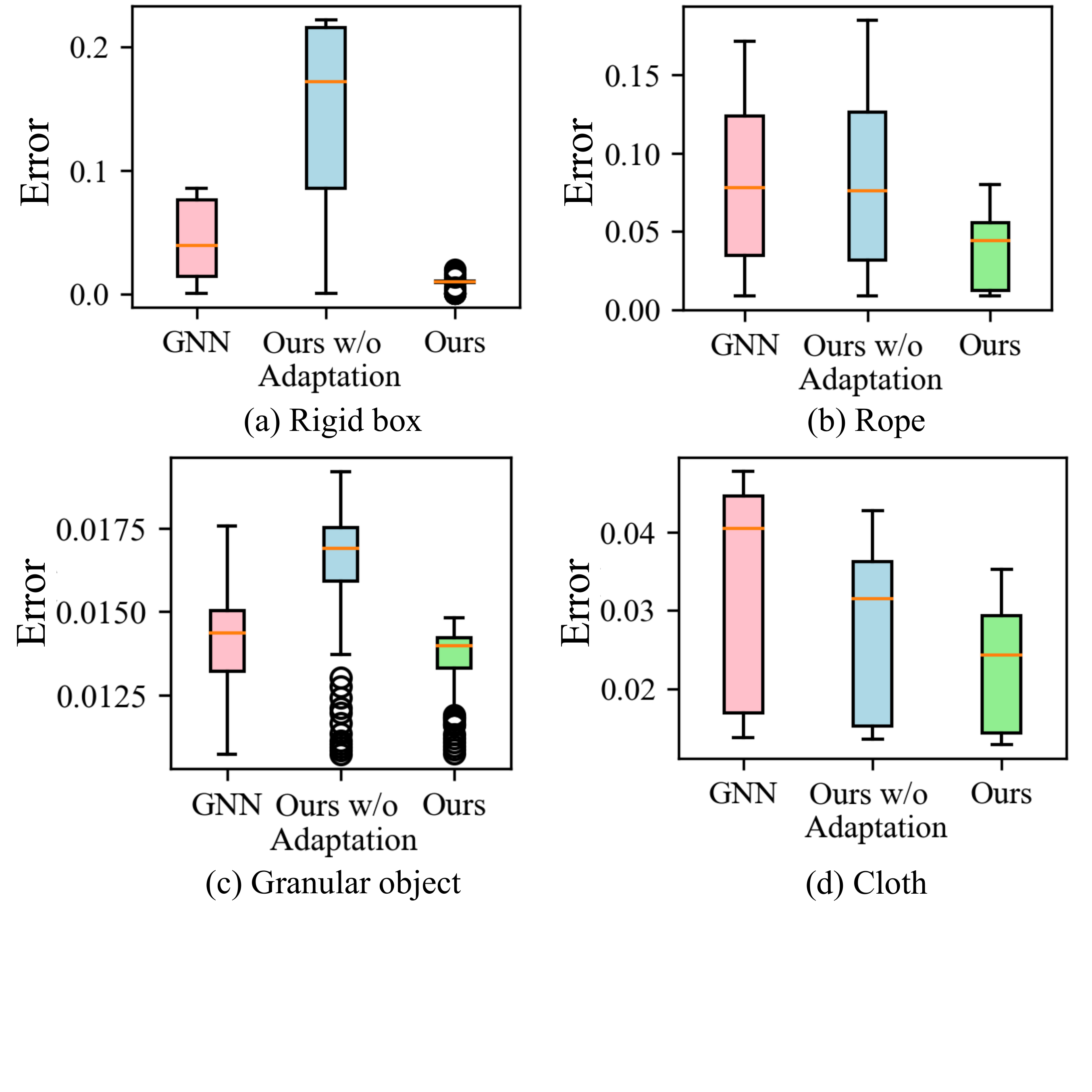}
    \vspace{-55pt}
    \caption{\small
    \textbf{Quantitative results on dynamics prediction.} We validate our model's effectiveness on a test set of 200 objects with distinct physical properties for each material type in simulation. Across all types of materials, our approach surpasses the baseline with respect to both the precision and consistency of predictions.
    }
    \label{fig:quant}
    \vspace{-20pt}
\end{figure}

\begin{figure*}[ht]
    \centering
    \vspace{-20pt}
    \includegraphics[width=\linewidth]{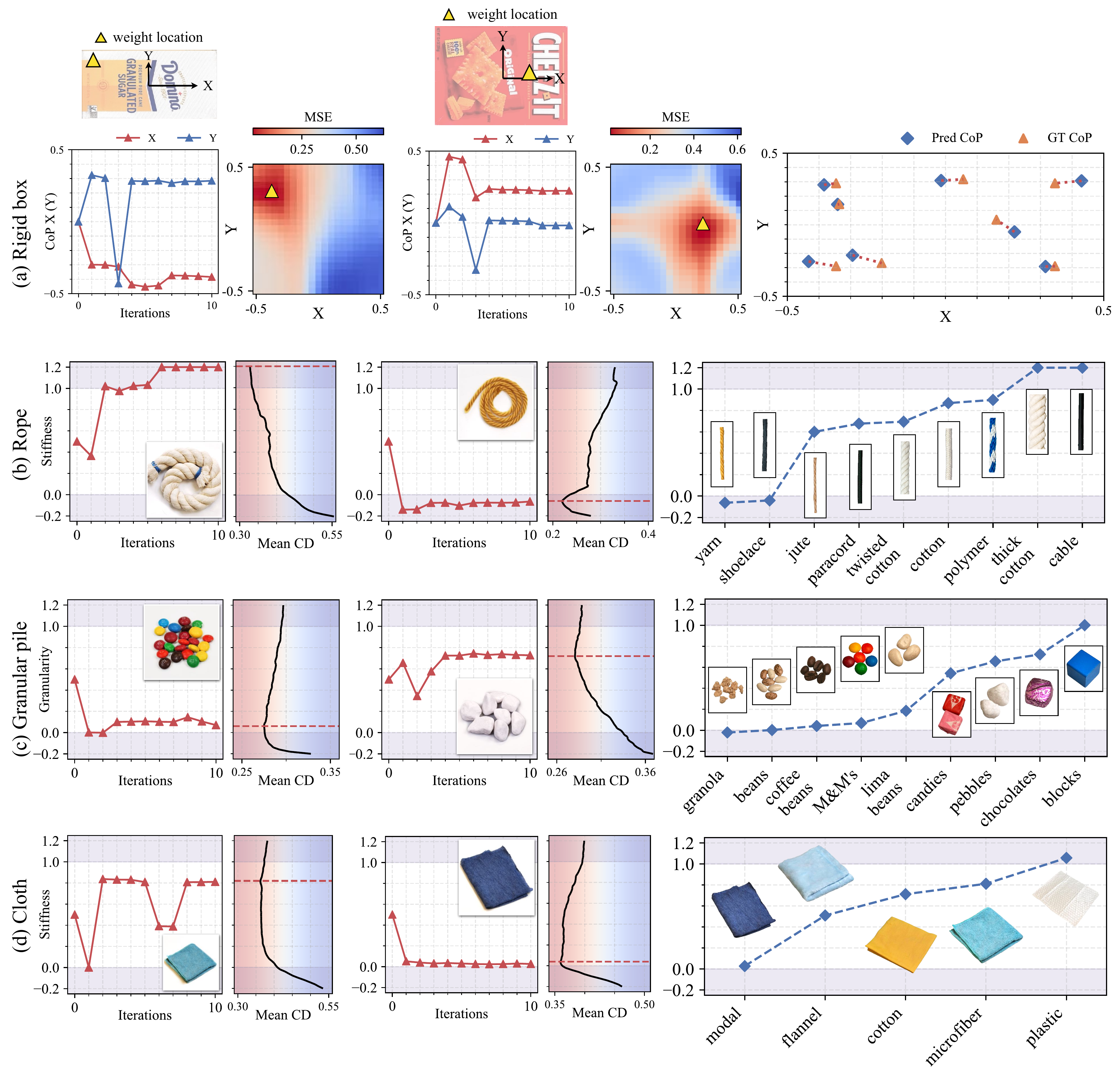}
    \vspace{-20pt}
    \caption{\small
    \textbf{Experimental results on physical property estimation.} Through the inverse optimization process, we estimate the physical properties of real-world objects. For each material type, we display the optimization trajectories alongside their associated costs, measured by the Chamfer Distance, for two objects with notably contrasting physical attributes. The rightmost column demonstrates that our estimated values align with human perceptions regarding the perceptual order of objects based on their physical property values, such as stiffness and granularity.
    }
    \label{fig:exp1}
    \vspace{-15pt}
\end{figure*}

\begin{figure*}[t]
    \centering
    \includegraphics[width=\linewidth]{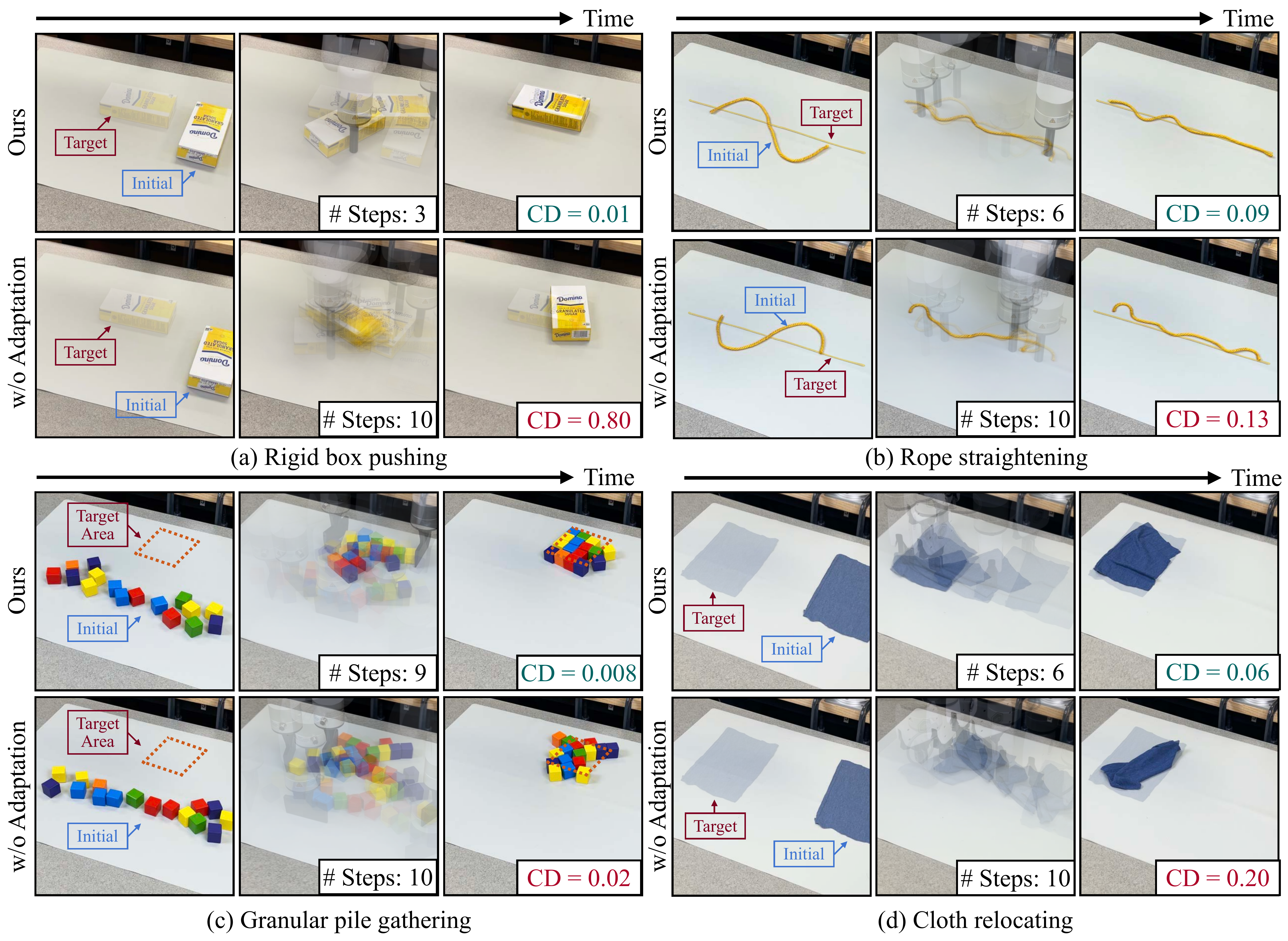}
    \vspace{-15pt}
    \caption{\small
    \textbf{Qualitative results on closed-loop feedback planning.} We present a qualitative comparison of MPC performance by contrasting our method across four tasks with the baseline model that does not employ physical property adaptation. Visualizations shown here demonstrate that our method effectively achieves the target configuration, whereas the baseline, even with more action steps, still exhibits a noticeable discrepancy compared to the target.
    }
    \label{fig:qual_control}
    \vspace{-18pt}
\end{figure*}
\begin{figure*}[ht]
    \centering
    
    \includegraphics[width=\linewidth]{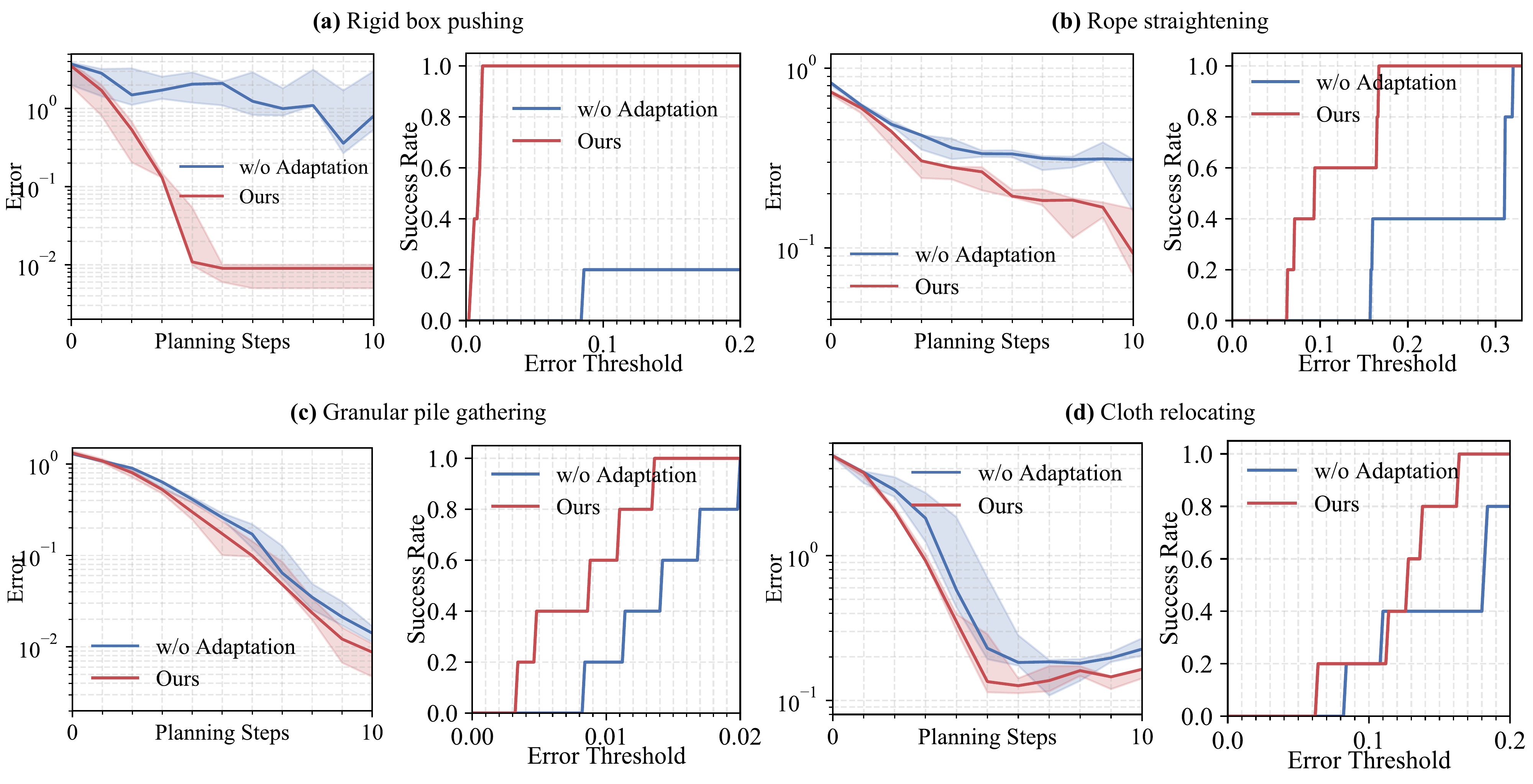}
    \vspace{-15pt}
    \caption{\small
    \textbf{Quantitative results on planning.} For each task, we use the same target configuration and initial configuration for the baseline method and our approach. We repeat each experiment-model pair 5 times and visualize (i) the median error curve w.r.t. planning steps (area between 25 and 75 percentiles are shaded) and (ii) the success rate curve w.r.t error thresholds. Our approach consistently outperforms the baseline method by being more accurate and using fewer action steps.
    }
    \label{fig:exp2}
    \vspace{-10pt}
\end{figure*}

In this section, we evaluate the proposed framework across a diverse range of object manipulation tasks. Our experiments are designed to answer the following questions: (1) Is the GBND model capable of accurately predicting the movements of objects with varied physical properties? (2) Can the test-time adaptation module effectively estimate real-world physical properties of objects through few-shot interactions? (3) To what extent does the integration of the adaptation module enhance the model's ability for model-based planning in the downstream manipulation tasks?

\subsection{Evaluation Materials and Corresponding Tasks}
\label{sec:tasks}

To demonstrate the modeling power of our framework for diverse materials, we implement one task for each of the 4 material categories: rigid box pushing, rope straightening, granular pile gathering, and cloth relocating.

\mypar{Rigid Box Pushing.}
The task is to use a point contact to push a box to a target position and orientation, which demands precise control over the translation and rotation motions in the presence of uncertainty of the center of pressure~\cite{zhou2019pushing}. The physical property variable is defined to be the normalized 2D position of the center of mass from the top view. As illustrated in Fig.~\ref{fig:overview}a, it is a 2-dimensional variable $\phi = [c_x, c_y]$ with range $c_x, c_y \in (-0.5, 0.5)$. We use the mean squared error as the cost function.

\mypar{Rope Straightening.}
The task is to rearrange the rope to a target configuration on the tabletop. We consider the stiffness of the rope as the physical property variable and define it as a normalized continuous variable $\phi \in (0, 1)$ where $\phi=0$ and $\phi=1$ correspond to the minimal and maximal stiffness in the simulator, respectively. 

\mypar{Granular Pile Gathering.}
The target is defined as a region on the tabletop, and the task is to gather the granular piles in an arbitrary initial distribution into the target region. We consider the granular size/granularity as the physical property variable and use a normalized variable $\phi \in (0, 1)$ to represent the size of a single grain in the pile.

\mypar{Cloth Relocating.}
The task is to use grippers to grasp the cloth and drag it on the table to place the cloth in the target configuration. We use a continuous variable $\phi \in (0, 1)$ to represent the stiffness of the cloth, which affects whether a piece of cloth will wrinkle or fold during a drag.

\subsection{Environment and Evaluation Setup} %
\label{sec:setup}

\mypar{Simulation.} 
Simulations of deformable and granular materials are conducted using NVIDIA FleX~\cite{li2018learning, macklin2014unified}, a position-based simulation framework designed to model interactions between objects of varying materials across multiple tasks, including pushing granular objects~\cite{Wang-RSS-23}, straightening ropes~\cite{corl2020softgym}, and unfolding clothing~\cite{ha2021flingbot}. Additionally, Pymunk~\cite{blomqvist2022pymunk} is utilized for simulating boxes that vary in shape and center of pressure.

For each material type, a dataset consisting of 1000 episodes is generated, with each episode featuring 5 random robot-object interactions. Within each episode, an object is assigned random physical properties (such as stiffness and granule size) that fall within a pre-defined range. To simulate interactions between the robot and the object, five random trajectories, involving either pushing or pulling actions, are created for every object. Throughout these interactions, data on the positions of particles and the robot's end-effector are gathered, which are then utilized for model training. More details on the simulation environment and data collection can be found in Sec.~\ref{sec:supp_sim} of the supplementary material.

\mypar{Real World.}
Fig.~\ref{fig:setup} presents the general setup in both the simulator and the real world. In the real-world experiments, we use a UFACTORY xArm 6 robot with 6 DoF and xArm's parallel gripper. For rigid box pushing and rope straightening tasks, we substitute the original grippers with a cylinder stick while we utilize a flat pusher for the granular pile manipulation task. These tools are 3D-printed and the same with the simulation setup to mitigate the sim-to-real gap. We fix four calibrated RealSense D455 RGBD cameras at four locations surrounding the workspace to capture the RGBD images at 15Hz and 1280x720 resolution. The robot manipulates objects within a 70\,cm$\times$45\,cm planar workspace. 

\mypar{Implementation Details.}
In all experiments, we assume the material type $M$ to be known, and the particles of the same object share the same $M$ and physical property variable $\phi$. To extract object point clouds from raw RGB-D inputs, we deploy the GroundingDINO~\cite{liu2023grounding} and Segment Anything~\cite{kirillov2023segment} model to detect and segment the table surface and objects. For the target object, we fuse the segmented partial point cloud from 4 views and apply a farthest point sampling method to a fixed pointwise distance threshold. For the cylinder stick and gripper, we use one particle to represent the end effector position, and for the flat granular pusher, we use 5 points to represent the end effector position and geometry.

\mypar{Baselines.}
To demonstrate the importance of parameter conditioning, we consider two baseline methods in our main experiments: (1) \textit{GNN} uses a graph neural network with the same architecture of our model, which is trained separately for each material category, but not conditioned on the physics parameter $\phi$. (2) \textit{Ours w/o Adaptation} is an ablated version of our material-adaptive model by using only the mean physical property variable $\bar{\phi}$ as input in deployment. \textcolor{black}{In Sec.~\ref{sec:supp_1} of the appendix, we include additional comparisons by finetuning the GNN baseline and adapting a physics-based simulator to demonstrate the effectiveness of our proposed conditioning method and the benefits of learned dynamics models.}

\subsection{Forward Dynamics Prediction}

Fig.~\ref{fig:qual_dynamics} shows the qualitative comparisons between our material-conditioned GBND model and the baseline model \textit{Ours w/o Adaptation}. The comparisons reveal that, with estimated physical property, the model's prediction matches the interaction outcome more accurately. For instance, in the rope scenario, the baseline model's prediction fails to capture both the below-average stiffness of the yarn object and the above-average stiffness of a polymer rope. In contrast, our method successfully accounts for variations in their motions, exhibiting more precise forecasts of unusual behaviors. Likewise, our model surpasses the baseline in scenarios involving materials with extreme physical properties, such as rigid boxes that differ in center of pressure, granular materials of various sizes, and clothes of differing stiffness.

Fig.~\ref{fig:quant} further validates our model's effectiveness on a simulated test set of 200 objects each with distinct physical properties. Our approach surpasses both baselines, \textit{GNN} and \textit{Ours w/o Adaptation}, demonstrating superior accuracy and stability for all the material types addressed in our study. Particularly, for rigid boxes, our model significantly outperforms the baselines with a near-perfect prediction accuracy.

\subsection{Physical Property Estimation}
For physical property estimation, we randomly initialize the object location on the tabletop and perform 10 interactions. 

\mypar{Rigid Box.} We use two boxes with different sizes: the sugar box (175mm$\times$89mm) and the cracker box (210mm$\times$158mm). We initialize the center of pressure (CoP) to be at 4 different locations for each box by putting weights at different locations inside the box. A visualization of all CoPs' normalized positions and our predicted CoP positions is shown in Fig.~\ref{fig:exp1}a. From the figure, we can observe that for all 8 data points, the predicted CoP positions are close to the ground truth CoP position. Moreover, the heatmap error shows that the low-error region for the CoP location forms a single global minima, and the predicted CoP positions converge to around the minimum value after around 5 interaction steps.  

\mypar{Rope.} We test our model on 9 different types of ropes. As shown in Fig.~\ref{fig:exp1}b, the model can extrapolate beyond the training data range [0.0, 1.0] and estimate out-of-range values for ropes with extreme stiffness/softness. The mean CD on the interaction observations gives clear and unique minimum points, and the stiffness ranking of the different types of ropes is consistent with the actual stiffness from human perception.

\mypar{Granular.} As shown in Fig.~\ref{fig:exp1}c, we test our model on 9 different types of granular objects by selecting representative objects of each granularity level,  ranging from approximately 1cm to 3cm. Results show that the predicted granularity ranking is consistent with the actual granular size. The model correctly predicts granola as the smallest grains and the toy blocks as the largest grains.

\mypar{Cloth.} As shown in Fig.~\ref{fig:exp1}d, we test our model on 5 different cloth instances, each with a different fabric material. The model correctly identifies the modal as the softest cloth (lowest stiffness). As another soft material, the flannel cloth is also estimated to be softer than cotton and microfiber cloths. While the training dataset does not contain any plastic-like materials, the model generalizes to a piece of plastic sheet and correctly predicts that it is very stiff.

\textcolor{black}{Furthermore, in Sec.~\ref{sec:supp_multipp} of the supplementary material, we present additional experiments that consider multiple parameters, namely the stiffness and friction of ropes. The results demonstrate that our method can be extended to recover more than one type of physical property simultaneously and yield better accuracy in dynamics prediction. }

\subsection{Model-Based Planning}

We further demonstrate that our material-conditioned GBND model and physical property adaptation can be integrated into an MPC framework to achieve a series of robotic manipulation tasks. Our experiments cover 4 distinct tasks outlined in Sec.~\ref{sec:tasks}, with a maximum limit of 10 planning steps imposed. Across all material types, our approach consistently meets the objectives within the allotted planning steps, unlike the baseline approach \textit{Ours w/o Adaptation}, which fails to achieve the goals due to its disregard for physical properties. For instance, in the rigid box pushing task, the baseline method incorrectly assumes the geometric center as the center of pressure, leading to inaccurate predictions of the box's straightforward movement post-push. Conversely, our method dynamically adjusts the center of pressure estimations during the interactions, thereby reaching the desired configuration in just three steps. Furthermore, as depicted in Fig.~\ref{fig:teaser}, the dynamics of pushing granular objects of different sizes vary significantly - larger granules push forward while smaller ones tend to stack and leave a trail. The baseline method, treating the motion of toy blocks and average granular piles similarly, fails to accumulate them in the target zone. Our method, however, identifies and adapts to the varied dynamics of granular materials, successfully completing the task.

Fig.~\ref{fig:exp2} offers quantitative results comparing the performance of our method against the baseline method \textit{Ours w/o Adaptation}, focusing on efficiency and error tolerance. Across four distinct tasks, our approach demonstrates superior performance, achieving lower errors within a constrained number of planning steps and attaining a higher success rate under a stringent error margin.

\section{Conclusion and Future Work}
\label{sec:conclusion}

We present AdaptiGraph, a unified graph-based neural dynamics framework for modeling multiple materials with unknown physical properties. We propose to condition the dynamics model on physical property variables and perform online few-shot physical property estimation. Experiments show that AdaptiGraph can precisely simulate the dynamics of multiple deformable materials, and adapt to objects with varying physical properties during deployment. We demonstrate the effectiveness of our framework across a wide range of objects in manipulation tasks. 

AdaptiGraph is a flexible framework. Currently, we train our model on four material types (ropes, granular objects, rigid boxes, and cloth) \textcolor{black}{and a single type of physical property for each material}. A future direction of our work is to extend our method to include more object materials and a more comprehensive set of physical properties that determine object dynamics. It is also possible to model heterogeneous object interactions using our framework by learning the dynamics model on a material-conditioned heterogeneous graph.

\section*{Acknowledgment}
This work is supported, in part, by NIFA Award 2021-67021-34418. We thank Mingtong Zhang, Binghao Huang, Yixuan Wang, and Hanxiao Jiang for the helpful discussions.

\bibliographystyle{plainnat}
\bibliography{references}

\clearpage
\begin{center}
\textbf{\textsc{\Large Appendix}}
\end{center}

\newcommand\DoToC{%
    \hypersetup{linkcolor=black}
  \startcontents
  \printcontents{}{1}{\textbf{Contents}\vskip3pt\hrule\vskip3pt}
  \vskip7pt\hrule\vskip3pt
}

\begin{appendices}

\renewcommand{\thesection}{\Alph{section}} 
\renewcommand{\sectionname}{}

\renewcommand{\thesubsection}{\Alph{section}.\arabic{subsection}}

\vspace{5pt}
\DoToC
\vspace{10pt}

\hypersetup{linkcolor=red}
\section{Comparison with Additional Baselines}
\label{sec:supp_1}

\subsection{Ablation on Material Conditioning}
\label{sec:supp_gnn}

Expanding on Fig.~\ref{fig:quant} from the main paper, we introduce an additional baseline, \textit{Unified GNN}, to study the importance of material type conditioning. As outlined in Tab.~\ref{table:baseline}, we establish the following baselines: (1) \textit{Unified GNN}, a singular GNN model trained on a combined dataset of rope, cloth, and granular materials, without any conditioning on material types or physical properties; (2) \textit{Separate GNN}, which employs a graph neural network with the same architecture as our model but lacks conditioning on physical parameters, and is independently trained for each material category; (3) \textit{Ours w/o Adaptation}, an ablated version of our material-adaptive model conditioned on the mean physical property variable $\bar{\phi}$.

\begin{figure}[t]
    \centering
    \hbox{
    \hspace{-10pt}
    \includegraphics[width=\linewidth]{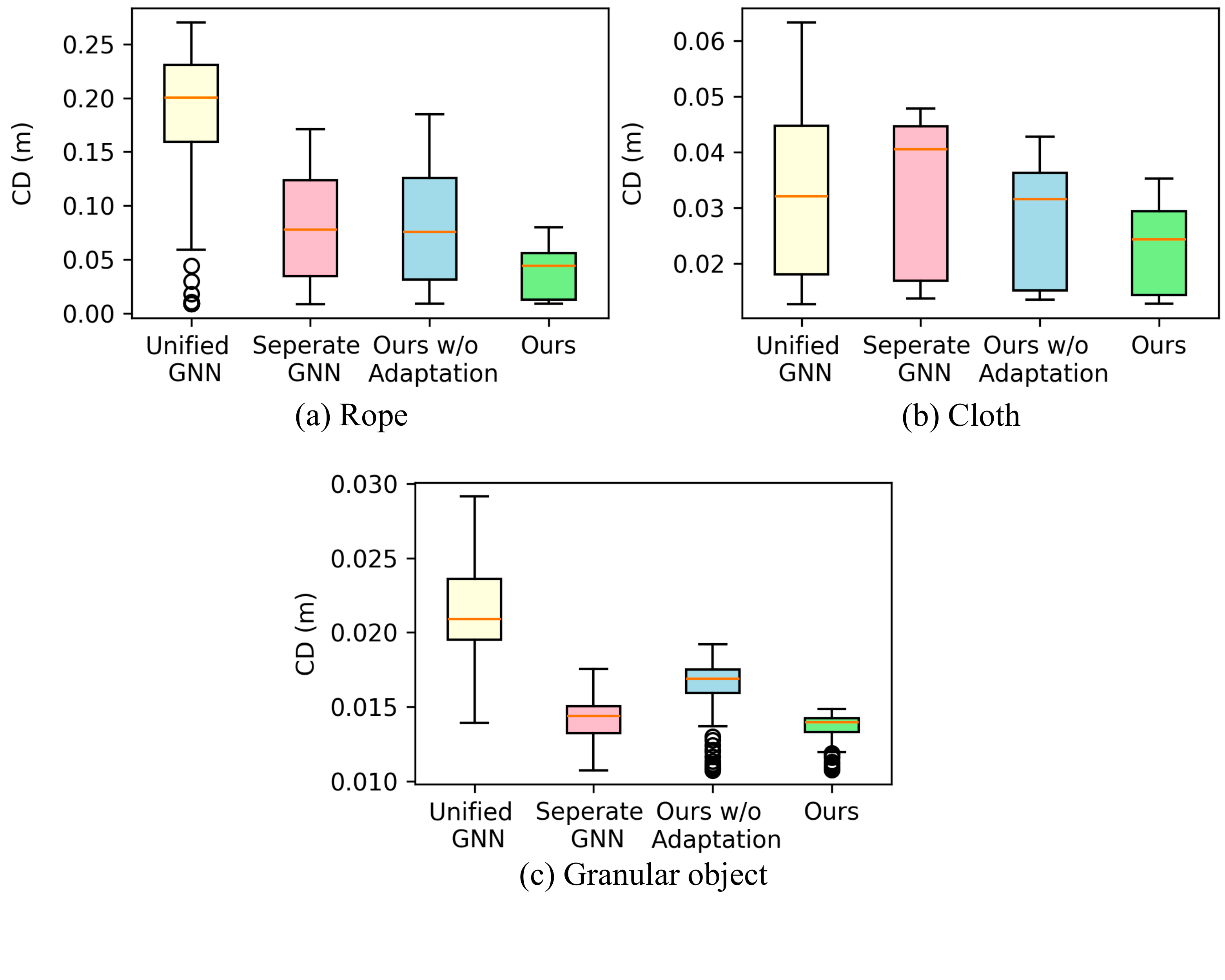}}
    \vspace{-20pt}
    \caption{\small
    \textbf{Quantitative results for ablation study.} We assessed our method and the baselines, as outlined in Tab.~\ref{table:baseline}, over 200 objects possessing diverse physical parameters across rope, cloth, and granular materials. 
    }
    \label{fig:baseline}
\end{figure}

The quantitative findings are displayed in Fig.~\ref{fig:baseline}. Within this assessment, the \textit{Unified GNN} has the lowest performance, with the highest variance, showing that it fails to model the complex dynamics brought by distinct material types and physical properties. The \textit{Seperate GNN} baseline performs better on individual material types than the \textit{Unified GNN}, reaching comparable performance with \textit{Ours w/o Adaptation}. However, the lack of physical property adaptability has led to inaccuracies. Overall, \textit{Ours} has achieved the best performance in all material types. The results have demonstrated the relative importance of material conditioning, physical property conditioning, and online adaptation in dynamics prediction performance.

\subsection{Adaptation Using Different Base Models}
\label{sec:supp_base_ada}

In our main paper, we have showcased the superior performance of our model in terms of dynamics prediction error when compared to two baseline models: \textit{GNN}, which does not use physical property conditioning, and \textit{Ours w/o Adaptation}, where the online adaptation module is removed. In this section, we further compare our model to (1) simulators incorporating physical property adaptation and (2) fine-tuning unconditional GNN.  We evaluate their dynamics prediction error after adaptation in few-shot real-world interactions.

\mypar{Settings.}
We employ the same simulators used for generating our training data: FleX~\cite{li2018learning, macklin2014unified} for deformable objects and Pymunk~\cite{blomqvist2022pymunk} for rigid boxes. Given the observed point cloud, we map the point cloud to object states in simulation with a perception model and then use the simulators to perform dynamics rollout and optimization-based physical property estimation. 

We design category-specific perception models to mitigate the sim-to-real gap. For boxes, we extract the 4 corners of the box from the top view and create an identical 2D box in Pymunk. For ropes, we apply mesh reconstruction based on alpha shapes~\cite{edelsbrunner1994three} to derive the rope mesh in the FleX simulator. For clothes and granular objects, we extract the contour of the point cloud's projection on the table surface, and construct object instances, i.e., a piece of cloth or granular pieces, that exactly cover the contour region.

\begin{table}[t]
    \centering
    \hbox{
    \hspace{-8pt}
    \begin{tabular}{ccccc}
    \toprule
    Method   & \begin{tabular}[c]{@{}c@{}}Unified \\ GNN\end{tabular}  & \begin{tabular}[c]{@{}c@{}}Separate \\ GNN\end{tabular} & \begin{tabular}[c]{@{}c@{}}Ours w/o \\ Adaptation\end{tabular} & Ours \\ \midrule
    Cond. on material type?         & $\usym{2718}$ & $\usym{2714}$                                                  & $\usym{2714}$                                           & $\usym{2714}$                                          \\
    Cond. on physical property? & $\usym{2718}$ & $\usym{2718}$                                                  & $\usym{2714}$                                           & $\usym{2714}$                                          \\
    Online adaptation?           & $\usym{2718}$ & $\usym{2718}$                                                  & $\usym{2718}$                                           & $\usym{2714}$                                          \\ \bottomrule
    \end{tabular}}
    \caption{\small
    \textbf{Difference in baseline models.} We ablate the online adaptation, material type conditioning, and continuous physical property conditioning modules in the listed baselines. The quantitative results are shown in Fig.~\ref{fig:baseline}.
    }
    \label{table:baseline}
\end{table}

\begin{figure*}[ht]
    \centering
    \includegraphics[width=\linewidth]{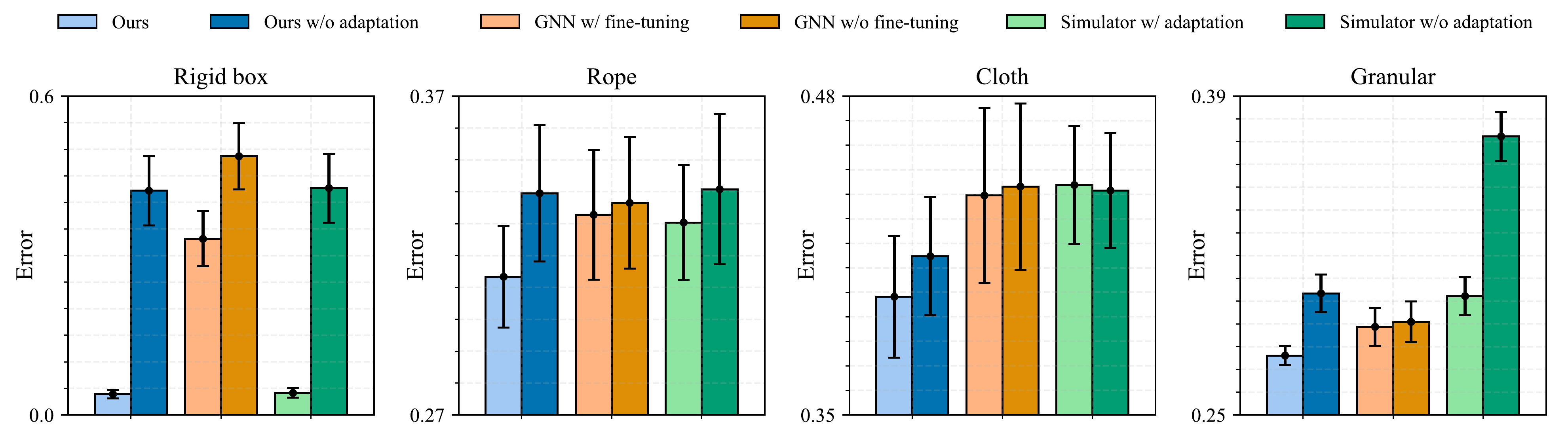}

    \caption{\small
    \textbf{Quantitative comparison with baselines adaptation approaches.} We report the mean and standard error of the dynamics prediction errors after online adaptation on 5 interactions. The numbers denote the mean value of each bar. For both material categories, our method achieves the lowest error across all methods after adaptation. Error metrics: rigid box - MSE, others - CD.
    }
    \label{fig:supp_exp1}
\end{figure*}

\mypar{Results.}
From Fig.~\ref{fig:supp_exp1}, we can observe that our method, with online adaptation, exhibits the lowest dynamics prediction error. It achieves an error reduction of $90.8\%$ for rigid boxes, $7.9\%$ for ropes, $4.0\%$ for clothes, and $9.0\%$ for granular objects compared to our model without adaptation. Notably, the error reduction ratio surpasses that achieved by fine-tuning an unconditional GNN-based dynamics model. Compared with simulator-based physical property adaptation, our model demonstrates $6.0\%$ lower dynamics prediction error for rigid boxes, $5.2\%$ for ropes, $10.3\%$ for clothes, and $8.7\%$ for granular objects. We attribute this improvement to the inherent system identification error and the instability of the simulator. Using a learning-based dynamics model directly on point clouds enhances our model's robustness to noisy visual inputs.

Moreover, our model is significantly faster than simulators. Running the Bayesian optimization algorithm for 50 iterations takes approximately 7 seconds for our model on a desktop computer equipped with an i9-13900K CPU and an NVIDIA GeForce RTX 4090 GPU, whereas it takes approximately 900 seconds for the FleX simulator.

\begin{figure*}[t]
    \centering
    \vspace{10pt}
    \includegraphics[width=\linewidth]{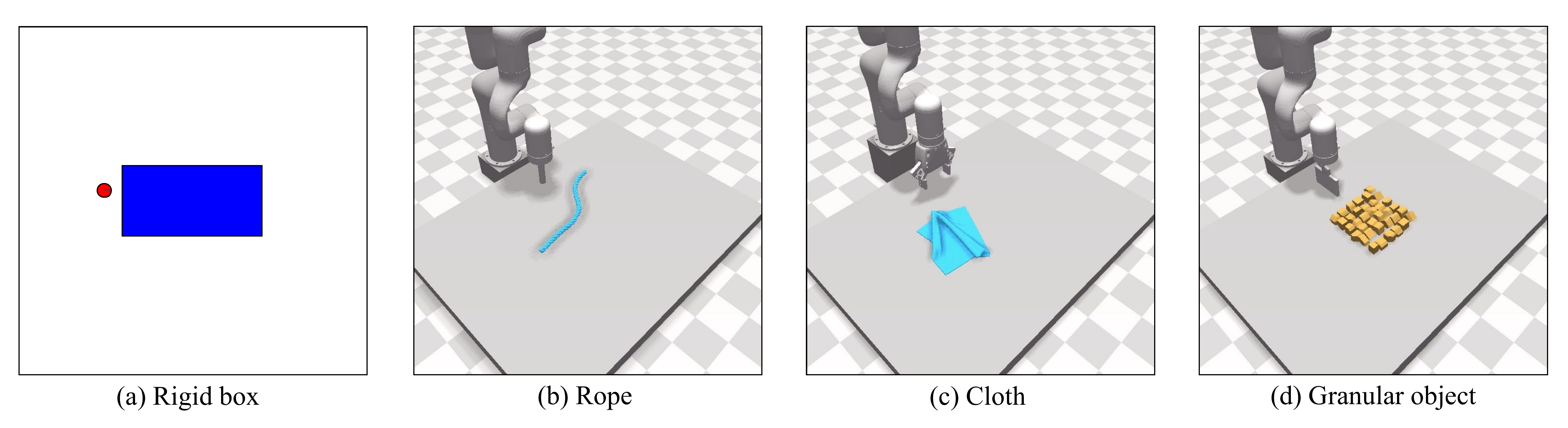}
    \caption{\small
    \textbf{Simulation environment visualization.} NVIDIA FleX is employed to create simulations of environments featuring ropes, cloths, and granular objects, incorporating a robot for interactions within a unified workspace. Additionally, we utilize Pymunk for simulating environments for rigid boxes with varying center of pressure locations.
    }
    \label{fig:simEnv}
\end{figure*}

\section{Additional Implementation Details}
\subsection{Simulation and Data Collection}
\label{sec:supp_sim}

For training our GBND model, we generate datasets encompassing variable physical properties in simulators. In FleX~\cite{li2018learning, macklin2014unified}, we render the robot workspace and the xArm6 robot mesh through OpenGL~\cite{opengl} from four camera angles to closely mirror our real-world configuration. In Pymunk~\cite{blomqvist2022pymunk}, the robot pusher is represented as a circular shape with a radius of 1cm. Fig.~\ref{fig:simEnv} shows the simulation setup for our data collection. The subsequent paragraphs detail the data generation process for each material type in the simulation. 

\mypar{Rigid Box.} As shown in Fig.~\ref{fig:simEnv}a, a circular rigid pusher interacts with a rigid box from random positions and angles. The length of the rigid box is uniformly sampled from 150$\sim$300 mm and the width is uniformly sampled from 50$\sim$200 mm. The center of pressure (CoP), represented by a 2-dimensional normalized coordinate in $(-0.5, 0.5)^2$, is sampled uniformly over the box surface. The friction coefficient between the box and the table is fixed as a control variable. We generate 1000 data episodes, each containing 1 pushing action on 1 box with random size and CoP.

\mypar{Rope.} As shown in Fig.~\ref{fig:simEnv}b, we use a cylinder pusher with a radius of 1cm to randomly interact with a simulated rope. The workspace in the simulation measures 90 cm$\times$70 cm. We uniformly randomize the length, thickness, and stiffness of the rope. We collect 1000 episodes of data, each comprising 5 continuous random pushes on one rope.

\mypar{Cloth.} The cloth simulation environment is created following the approach used in SoftGym~\cite{corl2020softgym}, as shown in Fig.~\ref{fig:simEnv}c. For the cloth properties, we vary the stretch stiffness (which determines resistance to elongation), bend stiffness (resistance to bending), and shear stiffness (resistance to sliding or twisting deformations). We randomly create rectangular clothes with lengths and widths uniformly distributed from 19 to 21 cm and 31 to 34 cm, respectively. We collect data across 1000 episodes, with each episode involving 5 continuous random interactions on one piece of cloth.

\mypar{Granular Object.} Adhering to the setup in~\cite{Wang-RSS-23}, we use irregular polygonal meshes to represent granular objects, as shown in Fig.~\ref{fig:simEnv}d. The scale of the granules is uniformly sampled in 1$\sim$3 cm. We also randomize the number of granular objects and the initial coverage area of the pile. The collected data consists of 1000 episodes, each comprising 5 continuous random interactions on one granular pile.

\subsection{Model-Based Planning} \label{sec:supp_mppi}

We apply the MPPI trajectory optimization algorithm for model-based planning. 
Given the dynamics model $z_{t+1} = f(z_t, u_t)$ (here we omit the material and physical property conditions for convenience), the cost function we minimize is:
\begin{equation}
    \mathcal{J}(u_{0:T-1}) = \phi(z_{T}, z^*) + l(z_0, u_{0:T-1}),
\end{equation}
where the task term $\phi(z_{T}, z^*)$ measures the distance from the current state to the target $z^*$, and the penalty term $l(z_0, u_{0:T-1})$ produces high cost for infeasible actions.

\mypar{Task term.}
For rope straightening and cloth relocating, the cost term is defined as the Chamfer Distance between the current state $z_T$ and the target state:
\begin{equation}
    \phi(z_{T}, z^*) = \text{CD}(z_T, z^*)
\end{equation}
For granular pile gathering, we use the nearest distance $\text{dist}_{z^*}(\cdot)$ from object particles to the target rectangle:
\begin{equation}
    \phi(z_{T}, z^*) = \frac{1}{|z_T|}\sum_{x\in z_T} \text{dist}_{z^*}(x),
\end{equation}
For rigid box pushing, since we have the correspondence between the observed box corners and the target corners, we use the Mean Squared Error (MSE):
\begin{equation}
    \phi(z_{T}, z^*) = \text{MSE}(z_T, z^*).
\end{equation}

\mypar{Penalty term.}
For all tasks, the penalty cost is defined as
\begin{align}\begin{split}
    l(z_0, u_{0:T-1}) &= \max_{x \in \mathcal{V}_T}\mathbbm{1}\{x \notin W\} \\
    &+ \max_{x_{\text{eef}}, x_{\text{obj}} \in \mathcal{V}_{0}} \mathbbm{1} \{\|x_{\text{eef}} - x_{\text{obj}}\| < d_{\min}\},
\end{split}\end{align}
where $W$ is the robot workspace; $\mathcal{V}_t$ is the particle set in state $z_t$; $x_{\text{eef}}$ and $x_{\text{obj}}$ represent end-effector and object particles, respectively. Thus, the penalty term penalizes actions that make the object particles move out of the workspace and the actions that will make the end-effector contact the object in $z_0$. We set $d_{\text{min}} = 2\text{cm}$ except for clothes where $d_{\text{min}} = 0$ as the grasping action allows contact.

\vspace{8pt}
\section{Discussion and Potential Extensions}

\begin{table}[]
\centering
\begin{tabular}{@{}ccccc@{}}
\toprule
                         & \multicolumn{2}{c}{Heuristics (Ours)}              & \multicolumn{2}{c}{Uncertainty-Driven}             \\ \cmidrule(l){2-5} 
\multirow{-2}{*}{Object} & {\color[HTML]{656565}  Estimation $\hat{\phi}$} & Variance $\sigma_{\pi_B}^2$    & {\color[HTML]{656565} Estimation $\hat{\phi}$} & Variance $\sigma_{\pi_B}^2$    \\ \midrule
Rope 1                   & {\color[HTML]{656565} 1.09}       & 0.029          & {\color[HTML]{656565} 1.20}       & \textbf{0.024} \\
Rope 2                   & {\color[HTML]{656565} -0.02}      & \textbf{0.025} & {\color[HTML]{656565} -0.01}      & 0.035          \\
Rope 3                   & {\color[HTML]{656565} -0.05}      & 0.030          & {\color[HTML]{656565} 0.00}       & \textbf{0.029} \\
Rope 4                   & {\color[HTML]{656565} 0.88}       & 0.020          & {\color[HTML]{656565} 0.83}       & \textbf{0.016} \\
Rope 5                   & {\color[HTML]{656565} 0.67}       & \textbf{0.018} & {\color[HTML]{656565} 0.63}       & 0.019          \\ \bottomrule
\end{tabular}
\caption{\small
\textbf{Uncertainty-driven identification results. } We show the estimated parameter $\hat{\phi}$ using both interaction selection methods for 10 interactions and the variance of the belief distribution $\sigma^2_{\pi_B}$ after optimization. Better results are in bold.
}
\label{table:uncertainty}
\end{table}

\vspace{2pt}
\subsection{Multiple Properties Recovering} 
\label{sec:supp_multipp}
\vspace{3pt}

Expanding on one-dimensional physical parameter conditioning for deformable objects, we designed an experiment to show that our method can also be applied to more than one physical property.

\begin{figure*}[t]
    \centering
    \vspace{10pt}
    \includegraphics[width=\linewidth]{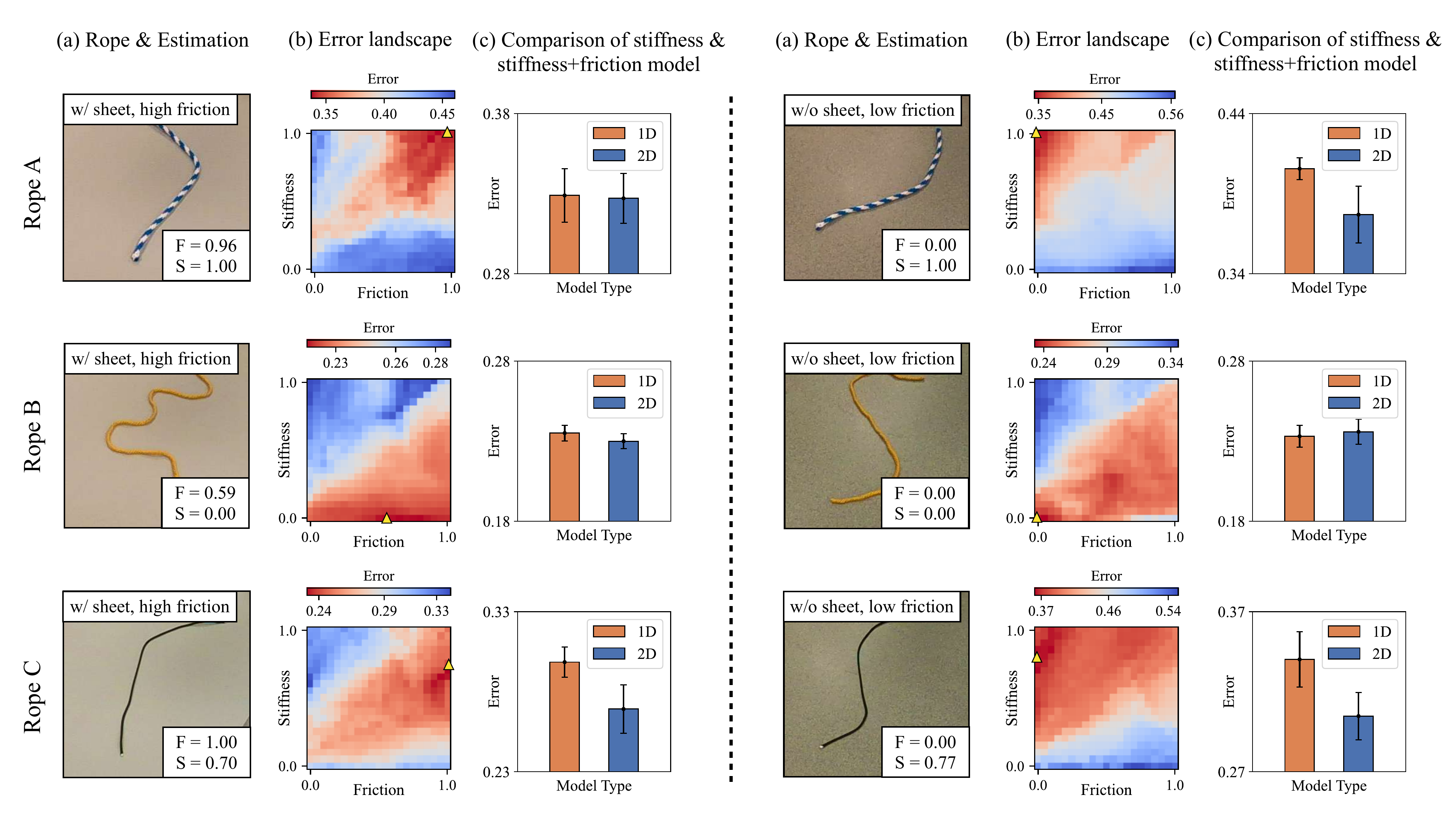}
    \caption{\small
    \textbf{Multiple physical properties estimations.} We test a total of 3 ropes (A: polymer, high stiffness; B: yarn, low stiffness; C: paracord, mid-level stiffness) on 2 different surface materials, \textit{w/ sheet} in which a rubber sheet is applied to increase friction, and \textit{w/o sheet} in which the rope is directly in contact with the table. For each combination, we show the estimated friction $F$ and stiffness $S$ (column a), the error landscape over the parameter space (column b, estimation results highlighted with yellow triangles), and the dynamics prediction error after adaptation, compared to the 1-dimensional stiffness-only model (column c).
    }
    \label{fig:multipp}
\end{figure*}

We consider ropes with 2-dimensional physical properties: stiffness $S$ and friction coefficient $F$. We generate the same amount of data as our previous setting in the Nvidia FleX~\cite{li2018learning, macklin2014unified} simulator with varying stiffness and friction and train a model conditioned on both properties. Then, we apply the model to ropes in the real world and perform property estimation and forward dynamics prediction. Results are shown in Fig.~\ref{fig:multipp}. As we can observe, the model can give reasonable estimates by predicting high friction in \textit{w/ sheet} cases and low friction in \textit{w/o sheet} cases. The stiffness estimations for all three ropes with and without sheets are also consistent. The dynamics prediction error for the 2D model (conditioned on both stiffness and friction) is generally lower than the 1D model (conditioned on stiffness only), showing that the dynamics prediction will be more accurate by incorporating more relevant properties.

\subsection{Identification with Uncertainty} \label{sec:supp_uncertainty}

The uncertainty of the physics parameters could be an important indicator measuring the estimation's confidence. Minimizing the uncertainty can also be used as an objective when selecting interactions~\cite{tremblay2023learning}. In comparison, our method selects actions that produce maximum displacement on object particles. In this experiment, we compare an uncertainty-driven interaction selection scheme with our heuristics-driven scheme.

We can define the belief state over the parameter space based on the dynamics prediction error:
\begin{equation}
    \pi_B(\phi) = \frac{1}{Z} \mathbb{E}_{(z_{i:i+1}, u_i) \in I} \left[e^{-\text{CD}(z_{i+1}, f(z_i, u_i; \phi, M)) / \tau}\right],
\end{equation}
where $\pi_B(\phi)$ is the probability density of physics parameter $\phi$ under belief $B$, $I$ is the set of interaction data, CD represents Chamfer Distance, $\tau$ is a temperature hyper-parameter which we set to $0.05$, and $Z$ is a normalizing factor. Parameters that give a lower dynamics prediction error will have higher probability density $\pi_B$. With this definition, a natural way to measure the uncertainty of a belief $B$ is by the uncertainty in the dynamics prediction outcomes, given current state $z$ and control action $u$:
\begin{equation}
    \mathbb{E}_{\phi, \phi' \sim \pi_B}\left[
    \text{CD}\left(f(z, u; \phi, M), f(z, u; \phi', M)\right)
    \right].
\end{equation}
Intuitively, by selecting an action $u$ that can maximize the uncertainty in the above equation, the interaction result will most effectively discriminate parameters sampled from the belief and thus be more effective in reducing the variance of $\pi_B$ post-adaptation. In practice, we sample $N$ parameters from $\pi_B$ and calculate the above equation as an MPPI objective.

Results of using this uncertainty-driven interaction selection approach are provided in Tab.~\ref{table:uncertainty}. We compare it with our heuristics-based approach and test on 5 different ropes. The estimated parameters $\hat{\phi}$ are consistent, and there is no consistent advantage in post-adaptation variance $\sigma^2_{\pi_B}$ over one another. Given that the heuristics-based approach is computationally faster, it is more suitable for our identification tasks.

\end{appendices}

\end{document}